\documentclass[10pt,twocolumn,letterpaper]{article}
\pdfoutput=1

\usepackage{cvpr}       

\usepackage{graphicx}
\usepackage{amsmath}
\usepackage{amssymb}
\usepackage{booktabs}
\usepackage{float}
\usepackage{algorithm}
\usepackage{algpseudocode}
\usepackage{tabularx}
\usepackage{makecell}
\usepackage[inline]{enumitem}

\usepackage[pagebackref,breaklinks,colorlinks]{hyperref}

\usepackage[capitalize]{cleveref}
\crefname{section}{Sec.}{Secs.}
\Crefname{section}{Section}{Sections}
\Crefname{table}{Table}{Tables}
\crefname{table}{Tab.}{Tabs.}

\algnewcommand\algorithmicforeach{\textbf{for each}}
\algdef{S}[FOR]{ForEach}[1]{\algorithmicforeach\ #1\ \algorithmicdo}

\def\cvprPaperID{3391} 
\def\confName{CVPR}
\def\confYear{2023}

\begin{document}

\title{A Generalization of Continuous Relaxation in Structured Pruning.}

\author{Brad Larson\\
Thermo Fisher Scientific\\
{\tt\small brad.larson1@thermofisher.com}
\and
Bishal Upadhyaya\\
Thermo Fisher Scientific\\
{\tt\small bishal.upadhyaya@thermofisher.com}
\and
Luke McDermott\\
UC San Diego\\
{\tt\small lmcdermo@ucsd.edu}
\and
Siddha Ganju\\  
NVIDIA Corporation\\
{\tt\small sganju@nvidia.com}
}
\maketitle

\begin{abstract}
Deep learning harnesses massive parallel floating-point processing to train and evaluate large neural networks. Trends indicate that deeper and larger neural networks with an increasing number of parameters achieve higher accuracy than smaller neural networks. This performance improvement, which often requires heavy compute for both training and evaluation, eventually needs to translate well to resource-constrained hardware for practical value. Structured pruning asserts that while large networks enable us to find solutions to complex computer vision problems, a smaller, computationally efficient sub-network can be derived from the large neural network that retains model accuracy but significantly improves computational efficiency.

We generalize structured pruning with algorithms for network augmentation, pruning, sub-network collapse and removal.  In addition, we demonstrate efficient and stable convergence up to 93\% sparsity and 95\% FLOPs reduction without loss of inference accuracy using with continuous relaxation matching or exceeding the state of the art for all structured pruning methods.  The  resulting CNN executes efficiently on GPU hardware without computationally expensive sparse matrix operations. We achieve this with routine automatable operations on classification and segmentation problems using CIFAR-10 \cite{CIFAR10}, ImageNet \cite{IMAGENET}, and CityScapes datasets \cite{CITYSCAPES} with the ResNet \cite{RESNET} and U-NET network architectures\cite{UNET}.

\end{abstract}

\section{Introduction}
\label{sec:intro}

\begin{figure}[ht!]
  \centering 
  \includegraphics[width=.50\textwidth]{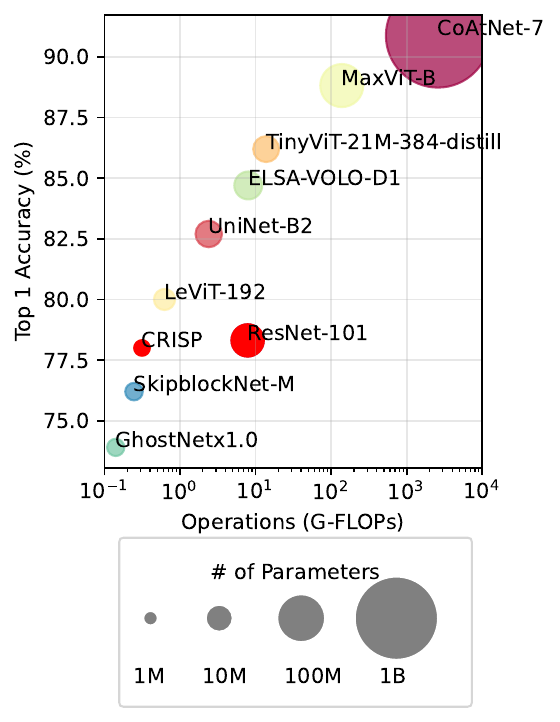}
  \caption{Classification results of various models trained on ImageNet-22K. 
  We (CRISP) maintain accuracy on ResNet-101 while decreasing total FLOPs with higher sparsity.}
\label{fig:ImageNet_FvA}
\end{figure}

An established trend in deep learning literature is the improvement of model accuracy with increased computation (see Figure \ref{fig:ImageNet_FvA}). This becomes challenging for model inference at the edge where data is abundant but compute resources are minimal due to power, hardware size, and cost constraints. Our focus is to reduce the inference computation of scientific and medical instruments, such as microscopes and CT scanners with structured pruning. For example, Transmission electron microscope (TEM) data can receive up to $4096X4096$ resolution images up to $400$ frames per second \cite{Falcon4i}, well past the achievable processing rates of large neural networks with the most advanced GPU. Structured pruning restructures models weights at training time.  A key assumption of pruning is that larger models provide great search spaces for pruning algorithms to find the subnetworks with optimal performance. The availability of large, transfer-learning models and their potential to be fine-tuned to certain tasks makes the structured pruning of larger models efficient.

Despite its published successes \cite{liu2021survey}, it is uncommon in practice to use structured pruning to improve model computational efficiency \cite{}.  Barriers to broadly adopting structured pruning as a standard part of CNN model optimization include (1) the need to manually customize models and model training \cite{DAIS:AutomaticChannelPruningviaDifferentiableAnnealingIndicatorSearch}, (2) published poor convergence stability \cite{}, and (3) unexplored behavioral differences between original and pruned models.  Here, we generalize the formulation of continuous relaxation in structured pruning into an automatable procure applicable to common convolutional neural network (CNN) architectures.  Where other structured pruning methods have surpassed the published performance of continuous relaxation in structured pruning, we simply and and improve the continuous relaxation in structured pruning formulation to achieve stable and direct convergence to state of the art structured pruning performance.  Finally, we evaluate the accuracy of image segmentation vs object size of original and pruned models.  This study demonstrates similar model behavior between original and pruned models, our desired behavior.  Structured pruning did not reduce intersection over union (IoU) accuracy irrespective of object size.  Our contributions include\footnote{All code, models, and example pipelines are available at \href{https://github.com/sherlj5352/crisp}{https://github.com/sherlj5352/crisp}.}:
    
\begin{enumerate}
    \item an automatable structured pruning and layer collapse algorithm applicable across CNN architectures,
    \item simplified stiffening and multi-step pruning providing direct and stable convergence to state-of-the-art (SOTA) computational efficiency without loss of accuracy,
    \item IoU vs object size metric demonstrating structured pruning maintained consistent model performance across of object sizes.
\end{enumerate}


\section{Related Work}

Neural network pruning methods can be divided into two categories – unstructured \cite{TheLotteryTicketHypothesis, AnOperatorTheoreticViewonPruningDeepNeuralNetworks, OnIterativeNeuralNetworkPruningReinitializationandtheSimilarityofMasks} and structured \cite{GaugingtheState-of-the-artforForesightWeightPruningonNeuralNetworks, GDP:StabilizedNeuralNetworkPruningviaGateswithDifferentiablePolarization, DAIS:AutomaticChannelPruningviaDifferentiableAnnealingIndicatorSearch}. Unstructured pruning removes parameters to optimize model sparsity. Specialized sparse matrix operations implemented in software or hardware are needed to realize a partial speedup from unstructured pruning.  Structured pruning provides a direct speedup with widely available GPU hardware and software deployments. Since its focus is computational efficiency, structured pruning removes entire network structures such as blocks, layers, and convolutional channels that maintain full matrix operations.
\cite{DAIS:AutomaticChannelPruningviaDifferentiableAnnealingIndicatorSearch}.

Structured pruning literature generally optimizes sparsity, in which the objective is to reduce the model size \cite{WinningtheLotteryAheadofTime:EfficientEarlyNetworkPruning}. Conversely, CNN floating point operation (FLOP) reduction applies greater weight to removing operations performed on larger channels that require more computations \cite{SNAP}.  FLOP reduction provides a hardware-independent metric of computational efficiency that can be quickly computed during model training. FLOP reduction is advantageous, especially for edge devices with numerous machine learning algorithms that vie for inference compute time. Flop reduction enables a greater batch size, reduced hardware requirements that free up space for instrument design, easier access to high-resolution data inference and a heightened efficiency of heterogeneous workflows.

We discuss related works along three threads:  
\begin{enumerate*}
    \item the achievement of high accuracy and sparsity through unstructured pruning and its impact on FLOPs,
    \item structured pruning search metrics, and regularization with sparsity approaching that of unstructured pruning and direct improvement in FLOPs,
    \item barriers to generalize and automate structured pruning.
\end{enumerate*}

The earliest works in neural network pruning show that randomly-initialized neural networks contain subnetworks that, when trained in isolation, achieve performance comparable to the original network \cite{TheLotteryTicketHypothesis}. To isolate the subnetworks, Lottery Ticket Rewinding (LTR) iteratively trains a network, prunes the smallest network parameters, \cite{WhentoPruneAPolicytowardsEarlyStructuralPruning}, and rewinds the remaining weights back to the original initialization. This and subsequent works \cite{EffectiveModelSparsificationbyScheduledGrow-and-PruneMethods, zhong2021revisit} demonstrate that greater than 97\% sparsity can be achieved through unstructured pruning on the classic CIFAR-10 benchmark of ResNet18. \cite{WinningtheLotteryAheadofTime:EfficientEarlyNetworkPruning}. The achievement of high sparsity exposed the bloated nature of large neural networks. However, their parameter removal did not translate into a proportionate improvement of computational efficiency \cite{kusupati2020soft}. Software acceleration of the sparse matrix operations have achieved a respectable fraction of the efficiency. \cite{gale2020sparse, sparsert, balancedsparsity}. Hardware-accelerated sparse matrix operations are likely to enable future performance improvements but also do not currently provide a speedup proportional to the sparsity achieved. \cite{SparsityNVIDIAAmpereTensorRT}.

To the aforementioned methods, structured pruning provides a simpler inference-time speedup and the ability to minimize computation (i.e. FLOPs) as an objective. The smallest element that can be pruned to accomplish this is dependent on the architecture.  For CNNs, the convolution channels are the most frequent unit for structured pruning as it provides sufficient granularity and search efficiency. \cite{GrowingEfficientDeepNetworksbyStructuredContinuousSparsification}. Earlier structured pruning approaches were inspired by unstructured pruning to identify the convolution channels to prune \cite{wen2016learning}. Then, the gradient \cite{WinningtheLotteryAheadofTime:EfficientEarlyNetworkPruning} and continuous relaxation methods grew to prominence. \cite{TrainingStructuredNeuralNetworksThroughManifoldIdentificationandVarianceReduction, GDP:StabilizedNeuralNetworkPruningviaGateswithDifferentiablePolarization}. Both approaches utilize gradient descent to select the pruned channels. The objective of gradient pruning, in short, is to remove channels with the smallest gradient adjustments. Continuous relaxation is our method of choice -- for which an additional relaxation parameter per convolutional channel is required.  The variable is used to continuously approximate the necessity of the channel. Continuous relaxation algorithms also include a method to progressively stiffen the approximation during optimization so the trained and relaxed model is akin to the pruned model \cite{DAIS:AutomaticChannelPruningviaDifferentiableAnnealingIndicatorSearch}.  With these approaches, structured pruning has been successfully applied to classification, semantic segmentation, and depth estimation with an upwards of 80\% FLOPs reduction while maintaining inference accuracy.

Structured pruning literature generally describe processes that require domain experts to prepare models for pruning \cite{GDP:StabilizedNeuralNetworkPruningviaGateswithDifferentiablePolarization}. It may also require an iterative train and prune cycle with a rigorous selection process for the pruned model. \cite{zhang2021learning}. We are developing a structured pruning algorithm that runs across a range of CNNs and does not require domain expert guidance, thus making it generalizable. We intend to integrate the pruning system within an automated model deployment pipeline in a production setting.

\cite{dong2019network}
\cite{tiwari2021chipnet}

\section{Method}

We first discuss the mechanics of network augmentation and pruning as a graph coloring problem. We then formalize the problem for CNNs and discuss how to identify which channels to prune. Finally, we devise a 
search strategy to minimize FLOPs while preserving accuracy.



 \begin{figure*}[ht!]  
  \centering 
  \includegraphics[width=.95\textwidth]{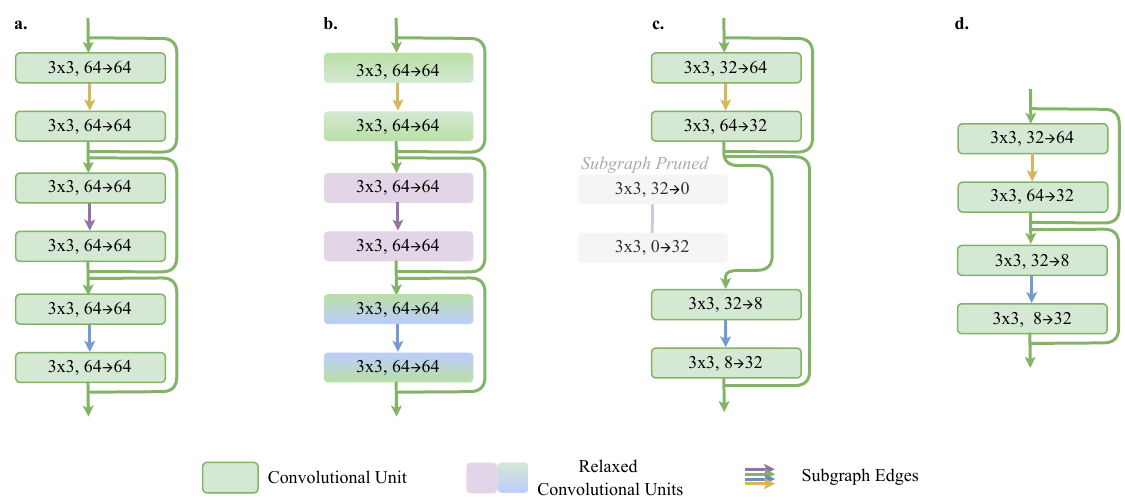}
  \caption{\textbf{CRISP method} \textbf{(a)} Directed acyclic graph representation of a ResNet pruning network. The graph edges (colored arrows) and their surrounding convolutions represent individual subgraphs. Throughout the pruning process, each subgraph is designed to maintain channel consistency between convolutional outputs and inputs. Note that the green subgraph calibrates the pruned channels of every residual unit. \textbf{(b)} Training and relaxation. The CNN weights are trained while the relaxation parameters learn the importance of individual convolutional channels. In this example, the golden sub-graph does not relax, the blue and green sub-graphs become partially relaxed and the violet sub-graph relaxes completely.  \textbf{(c)} Pruning. All convolutional channels with a relaxation parameter close to zero are pruned. The violet sub-graph is entirely pruned. \textbf{(d)} The final network after pruning and sub-graph collapse.}  
\label{fig:crisp-overview}
\end{figure*}

\begin{table*}
  \centering
\begin{tabular}{|c|c|c|}
\hline
Operator & Input & Output\\
\hline \hline
Concatenation $ \oplus $ & $T_{i,n} = (b,c_n)\oplus d $ & $T_o = (b, \sum c_n)\oplus d $\\
\hline
ReLu & $ T_i = (b,c)\oplus d $ & $ T_o = T_i $ \\
\hline 
Batch Norm & $T_i = (b,c)\oplus d $ & $ T_o = T_i $ \\
\hline 
Sum $u+v$ & \makecell{$T_{i,u} = (b,c)\oplus d$ \\ $T_{i,v} = (b,c)\oplus d$ }  & $ T_o =(b,c)\oplus d $ \\
\hline 
\makecell{Element-wise\\Product $ u \cdot v $ }  & \makecell{$T_{i,u} = (b,c)\oplus d$ \\ $T_{i,v} = (b,c)\oplus d$ } & $ T_o = (b,c)\oplus d $\\
\hline 
Convolution $ \circledast $ & $ T_i = (b,c_i)\oplus d_i $ & $ T_o = (b,c_o)\oplus d_o $ \\
\hline 
Fully Connected & $T_i = (b,c_i) $ & $ T_o = (b, c_o)$ \\
\hline
\end{tabular}
  \caption{Neural Network Operators with input and output dimension. $b$: batch size, $c$: number of channels, $d$: array of image spatial dimensions, $k$: channel mask vector, $m$: array of convolution kernel dimensions, $T$: tensor size vector}
  \label{tab:size}
\end{table*}

\subsection{Subgraph Identification and Augmentation}\label{Subgraph Identification and Augmentation}

To generalize the network augmentation, training, and pruning procedures, we consider the CNN to be a Directed Acyclic Graph (DAG) as illustrated in Figure \ref{fig:crisp-overview} (a). Graph nodes are the network operators and edges are the tensor outputs of one node and inputs go to the next node. Modeled on ResNet, each block in Figure \ref{fig:crisp-overview} (a) contains a 2D convolution, batch normalization, and ReLU operator.
Table \ref{tab:size} outlines the input and output tensor size for common neural network operators. Most operators preserve the size of the input. For example, a ReLU activation output is identical to its input, convolutional and fully connected layers are the exception. 

To ensure tensor shape consistency, we introduce sub-graphs to the DAG.  Each sub-graph is: \begin{enumerate*}
    \item bound by convolutional layers and,
    \item consists of operators that must necessarily maintain equivalent input and output shapes throughout the pruning process.
\end{enumerate*}
Figure \ref{fig:sparsecurve} (a) showcase the nuances of a ResNet DAG. Its subgraphs are represented by distinct colored edges and their associated convolutional units. The residual sub-graph spans across ResNet blocks as the element-wise addition of the identity function requires equivalent tensor shapes. Each sub-graphs are candidates for augmentation, relaxation and pruning. 

An augmented subgraph $f_a(x,w,\sigma(s) )$ is the product of the subgraph $f_s(x,w)$ with the relaxation function $\sigma(s)$ as shown in Equation\ref{equ:augmentedsubgraph}

\begin{equation} 
f_a(x,w,\sigma(s) ) = f_s(x,w) \cdot \sigma(s)
\label{equ:augmentedsubgraph}
\end{equation}

where the sigma function $\sigma(s)$ inputs the relaxation parameter $s \in \mathbb{R}$ and outputs a value from 0 to 1. The relaxation function enables a continuous optimizer to explore the discrete problem of enabling or disabling each relaxed channel as part of model training. To prune a neural network, the relaxation function is converted to a binary mask $m$ using a threshold value $\tau$ and $\sigma(s)$:
\begin{equation} \label{eq:Prune}
m = 
    \begin{cases}
        1 & \text{if } \sigma(s) > \tau\\
        0 & \text{else}
    \end{cases}
\end{equation}
The binary mask is applied to each operator within a sub-network retaining only the candidate channels.  After pruning, the unpruned relaxation function outputs $\sigma(s)$ can be retained as a channel bias or better, merged into adjacent operators reducing computation. If all channels are pruned then the entire sub-graph is removed. In addition, incoming and outgoing operators are removed from the graph until a non-zero path exists.

This procedure can be generalized to any deep learning architecture with at least two consecutive convolutional layers. Sub-graphs that contain unknown operators may be excluded from pruning. 

\subsection{Problem Formulation} \label{Problem Formulation}
We start with an accuracy objective that searches for convolution weights $w$ and relaxation weights $s$ to minimize the objective function $\mathcal{L}(f(x,w,\sigma(s)))$. 
Second, we add in the structural pruning objective, and scale it with an architecture weighting factor $\mu \| \sigma(s) - t \|_1$, which approaches zero as the model structure $\sigma(s)$ approaches the target structure $t$. The model structure is expressed as two forms:
\begin{equation}  
\sigma_p(s) = \frac{1}{P} \cdot \sum_{i} p_i \cdot \cfrac{1}{1+e^{-as}}
\end{equation}
\begin{equation}  
\sigma_q(s) =  \frac{1}{Q} \cdot \sum_{i} q_i \cdot \cfrac{1}{1+e^{-as}}
\end{equation}

Where $p_i$ is the operator parameters (see Supplementary Section), $P$ is the total model parameters, $q_i$ the operator FLOPs, $Q$ being the total model FLOPs, and $a$ as the logistic growth rate or steepness of curve.

The model structure is based on the logistic function. We formulate the logistic function to approach $0$ for small $s$ and $1$ for large $s$ providing a bounded, differentiable range $(0,1)$ from the continuous domain of $s \in \mathbb{R}$. $\sigma_p(s)$ weighs model parameters for comparison with other structured and unstructured pruning algorithms' sparsity whereas $\sigma_q(s)$ weighs model FLOPs to improve computational efficiency. 

Third, we add in the stiffening objective scaled by its weighting factor to ensure that the relaxed model before and after pruning is similar. This is achieved when the relaxation of the pruning weights is far from 0. 

We observe smoother convergence and improved pruning performance when model stiffening is an optimization objective. The stiffening function is:
\begin{equation}
\delta(s) = \frac{1}{N} \sum_{i} \exp\left({\frac{-s^2}{2b^2}}\right)
\label{eq:stiffining}
\end{equation}

This is derived from the Gaussian probability density function:
\begin{equation}
\exp\left({\frac{-s^2}{2b^2}}\right) \propto \mathcal{N}(s|0, b^2)
\label{eq:gaussian}
\end{equation}

where
\begin{itemize*}
\item[$N$]: \text{Number of pruning weights}
\item [$b$]: \text{Gaussian Standard Deviation}.
\end{itemize*}

which defines a Gaussian function for each of the pruning weights.

Thus our final objective function for a given CNN $f$, model weights $w$ and relaxation parameters $s$ becomes:
\begin{equation} \label{eq:1}
\mathcal{T}(x,w,s) = \mathcal{L}(f(x,w,\sigma(s))) + \mu \| \sigma(s) - t \|_1 + \lambda \cdot \delta(s)
\end{equation}
\begin{equation} \label{eq:TotalLoss}
\min_{w,s} \sum_{x \in X} \mathcal{T}(x,w,s)
\end{equation}
such that, \\
\begin{itemize*}
\item[$\mathcal{L}$]: \:\text{accuracy loss}, \\
\item[$\mathcal{T}$]: \:\text{total loss}, \\
\item[$f$]: \:\text{convolutional neural network}, \\
\item[$X$]: \:\text{set of input tensors}, \\
\item[$V$]: \:\text{set of ground truth label tensors}, \\
\item[$w$]: \:\text{network weights}, \\
\item[$s$]: \:\text{pruning weights}, \\
\item[$\sigma$]: \:\text{architecture minimization function}, \\
\item[$t$]: \:\text{target architecture}, \\
\item[$\mu$]: \:\text{architecture weighting factor}, \\
\item[$\|\cdot\|_1$]: \:\text{L1 Norm}, \\
\item[$\lambda$]: \:\text{stiffening weighting factor}, \\
\item[$\delta$]: \:\text{stiffening function}\\
\end{itemize*}

Since our objective is to minimize model size without reducing model accuracy, $\mu$ and $\lambda$ are chosen to be close to the minimum value of the objective function.  This results in an structured pruning without reducing model accuracy.


A high $\lambda$ at initialization inhibits sparsity because it enforces $s$ to stay on the same side as the critical point of our Gaussian stiffening function, $\delta$. However, waiting a few iterations in training allows the initial search to begin, and increasing $\lambda$ later induces ``confidence'' in our determined $s$-values as it pushes to positive or negative infinity. In layman's terms, we do not want the search to be confident of the initialized $s$-values. Increasing $\lambda$ later results in a faster convergence of $s$.

\subsection{Search}\label{search}
 
Algorithm \ref{CRISPalgorithm} formulates a reliable and efficient algorithm to search for the minimum FLOPs that preserves model accuracy. 

\begin{algorithm}
\caption{Continuous relaxation in structured pruning}\label{alg:structured pruning}
\begin{algorithmic}
\Require Training set $X$ and label set $V$
\State $f(x,w,\sigma(s))$ \Comment{network with subnetworks}
\State $w = w_0$ \Comment{initialize/transfer network weights}
\State $s = s_0$ \Comment{initialize/transfer pruning weights}
\State $W $ \Comment{workflow definition} 
\State $j = j_{max}$ \Comment{workflow steps}
\State $i = i_{max}$ \Comment{training epochs}
\While{$j \geq 0$}
  \If{$W_{j, prune}$}
    \ForEach {$s \in \mathcal S $} \Comment{For each subgraph}
        \State $m = \sigma(s) > \tau_j$  \Comment{Prune mask}
        \State $P(f_s(x_s,w_s,\sigma(s_s)))$  \Comment{Prune the subgraph}
    \EndFor
  \EndIf
  \If{$W_{j, train}$}
    \While{ $i \geq 0$ }
      \State $ \min_{w,s} \sum_{x \in X} \mathcal{T}(x,w,s) $ 
      \Comment{Train}
      \State $i=i-1$
    \EndWhile
  \EndIf
  \If{$W_{j, test}$}
    \State $r = \sum_{i} \displaystyle{\| f(x_i,w,\sigma(s) - v_i \|}$ \Comment{Test Accuracy}
  \EndIf
  \State $j=j-1$
\EndWhile
\State $f(x,w,\sigma(s))$ \Comment{network with pruned sub-networks}
\State $w$ \Comment{final trained weights}
\State $s$ \Comment{final pruning weights}
\State $r$ \Comment{test results}
\end{algorithmic}
\label{CRISPalgorithm}
\end{algorithm}

Since continuous relaxation evaluates channels as fractionally present, it is feasible to perform structured pruning in a single step \cite{yuan2020growing, DAIS:AutomaticChannelPruningviaDifferentiableAnnealingIndicatorSearch}.  However, Algorithm \ref{CRISPalgorithm} implements a workflow iteration "while {$j \geq 0$} do".  This addition to continuous relaxation is inspired by early pruning literature \cite{WhentoPruneAPolicytowardsEarlyStructuralPruning, WinningtheLotteryAheadofTime:EfficientEarlyNetworkPruning}.  Concerned about pruning weights early that become important only late in training, we introduced the workflow dependent $\tau_j$ with an initially small prune threshold (e.g. 0.01) that we ramp progressively towards 0.5. Experimentally, we have not achieved superior single-step pruning results to multi-step pruning following this approach.  Both converge towards a similar pruned structure with multi-step pruning having a smoother convergence.  In addition, early pruning speeds model training as the model size decreases. The workflow definition variable $W$ allows us to train and test on the first pass followed by prune, train, and test on subsequent passes.

\section{Experiments}
The premise of CRISP is to 1) train everyday neural networks into high accuracy while minimizing computation and 2) demonstrate that this can be achieved automatically in a model development pipeline. To establish this, we first determine the highest sparsity attainable for various pruning algorithms and benchmark CRISPs performance in comparison. We observe that all the models remain under the theoretical limit specified by LTR. Then we investigate the stability of the training convergence to determine the generalizability of the algorithm. Our data shows a repeatedly robust convergence when optimizing for standard cross-entropy loss, mean intersection-over-union (mIoU) and FLOPs. Finally, the intent of CRISP is to deploy into real-world systems where class objects vary by size and shape tremendously, especially due to data drift. To verify the performance of pruned models on complex class objects, we carried out a size-based analysis on the human class of the Cityscapes dataset. 

We perform our experiments on standard datasets to ensure a fair comparison to existing works so that future users can utilize the benchmarks to decide the best methods for their use cases. We use PyTorch framework for experiments on NVIDIA RTX 5000 and NVIDIA RTX 6000 for CIFAR10 datasets, and NVIDIA A6000 and NVIDIA A100 for CityScapes\cite{cordts2016cityscapes} and ImageNet dataset. All the parameters, models, and code are available at \href{https://github.com/sherlj5352/crisp}{https://github.com/sherlj5352/crisp}.

\subsection{Accuracy vs Sparsity Comparison}\label{Accuracy vs Sparsity Comparison}
\begin{figure}[ht!]  
  \centering 
  \includegraphics[]{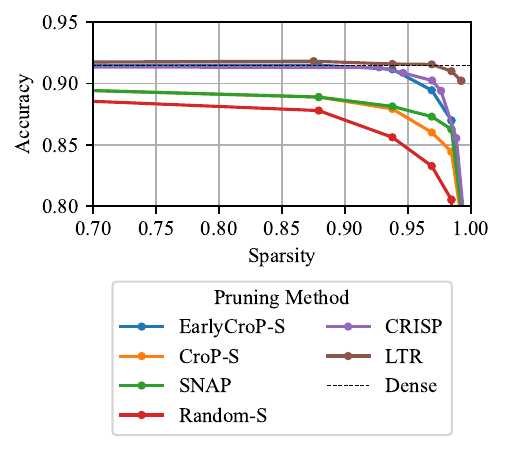}
  \caption{Model accuracy vs sparsity with structured pruning. Performance bounded by LTR unstructured pruning and random structured pruning.}
\label{fig:sparsecurve}
\end{figure}

First, we search for the highest sparsity attainable. With data provided by structured and unstructured pruning algorithms alike, Figure \ref{fig:sparsecurve} explores the trade-off between network accuracy and sparsity of the ResNet-18 networks on top-1 accuracy of the CIFAR-10 dataset. Although unstructured pruning achieves a lower speedup at the same sparsity, it provides a theoretical upper-limit for structured pruning sparsity. Random channel pruning (Random-S in the figure) provides the lower bound where channels are randomly pruned followed by iterative model retraining and testing. For all the pruning methods, we start from the same baseline performance of 91.5\% which is the off-the-shelf performance of the ResNet-18 model on CIFAR-10. 
Our goal is to maintain model accuracy (reduction of less than 1\%) as sparsity increases.

We note that existing methods and CRISP exhibit an elbow-shape as the accuracy plummets at high sparsity.  CRISP has its elbow point at 96.9\% sparsity with the accuracy preserved at 90.25\%, akin to the structured pruning state-of-the-art. We define the preservation of accuracy to be within a 1\% threshold of the off-the-shelf model.

Early-CroP-S\cite{WinningtheLotteryAheadofTime:EfficientEarlyNetworkPruning} (at 95\% sparsity and 91\% accuracy). At 94.6\% sparsity, CRISP performs with 90.87\% accuracy. 


\begin{center}
\begin{tabular}{c c c c} 
 \hline
 Method & Accuracy & Sparsity & \% FLOPS \\ [0.5ex] 
 \hline\hline
 PFEC [30] & 93.06\% & - & 27.6\%\\ 
 \hline
 LCCL [7] &  92.81\% & - &  37.9\% \\
 \hline
 AMC [15] &  91.90\% & - & 50.0\% \\
 \hline
 SFP [18] & 93.35\% & - & 52.6\% \\
 \hline
 FPGM [19] & 93.49\% & - &  52.6\% \\ 
  \hline
 TAS &  93.69\% & - & 52.7\% \\ 
 \hline
  PFS  &   93.05\% &  50 & 52.7\% \\ 
 \hline
  HRank \cite{lin2020hrank} &   93.17\% &  42.4\% & 50.0\% \\ 
 \hline
  ABCPruner \cite{lin2020channel} &   93.23\% &  - & 54.1\% \\ 
 \hline
  DAIS \cite{DAIS:AutomaticChannelPruningviaDifferentiableAnnealingIndicatorSearch} &   93.53\% &  - & 70.9\% \\ 
 \hline
\end{tabular}
  CIFAR-10 Resnet-56\cite{dong2019network,DAIS:AutomaticChannelPruningviaDifferentiableAnnealingIndicatorSearch}
  \label{tab:objective}
\end{center}

\subsection{Sparsity vs FLOPs}\label{Training Convergence Stability}

Whereas a sparsity objective treats all pruning weights equally, the FLOPs objective favors CNN channels that process and propagate data of larger resolutions. With the application of structured pruning on the U-NET \cite{DBLP:journals/corr/RonnebergerFB15} image segmentation of CityScapes\cite{CITYSCAPES} we seek to answer the following questions:  
\begin{itemize*}
\item[(a)] What is the trade-off in performance when optimizing for sparsity vs FLOPs? 
\item[(b)] What are the structural difference between models pruned for sparsity vs for FLOPs?
\end{itemize*}

\begin{figure}[ht!]  
  \centering 
\includegraphics[width=.45\textwidth]{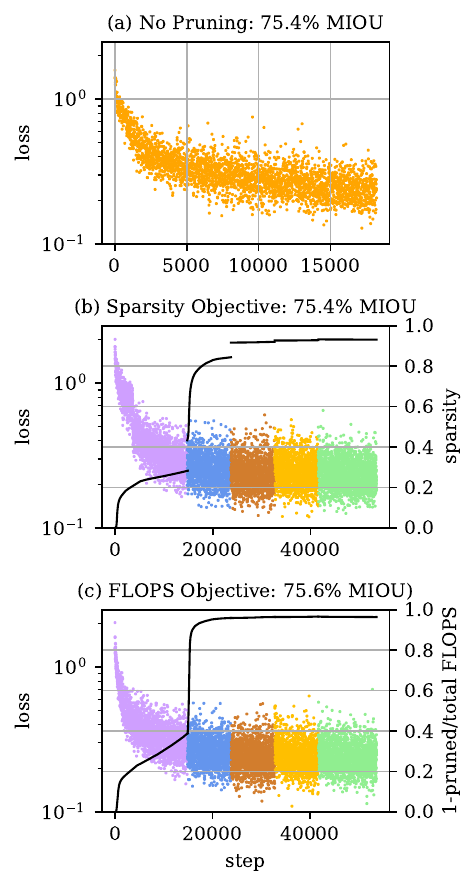}
  \caption{Training monitoring test loss and pruning ratio through (a) no pruning: 75.4\% mIoU, Params 31.04e6, 218.9 GFLOPs, (b) sparsity objective: 75.4\% mIoU, 91.0\% Sparsity, 33.7 GFLOPs, and (c) FLOPs objective: 75.6\% mIoU, 91.2\% Sparsity, 15.5 GFLOPs}
\label{fig:converge}
\end{figure}


Figure \ref{fig:converge} shows the training convergence of three training sequences: (a) no pruning: train only to minimize standard cross-entropy loss, (b) CRISP sparsity: a 5-step CRISP training using the sparsity objective, (c) CRISP FLOPs: a 5-step CRISP training using the FLOPs objective. The scatterplot shows the cross entropy loss of the validation dataset evaluated periodically during training on left Y-axis. The black line associated with the right Y-axis displays the sparsity in Figure \ref{fig:converge} (b) and the FLOPs removal in Figure \ref{fig:converge} (c). Each color represent one step of the iterative training and pruning workflow.

As outlined in Section \ref{search}, the first CRISP training step (purple scatterplot in Figure \ref{fig:converge} (b) and (c)), seeks to minimize cross entropy, architecture and stiffening losses.  Every subsequent step prunes the model, then resumes its training using the same loss objectives.  The cross-entropy loss tends to jump immediately after pruning especially at higher sparsity.  However, the shift is small and overall accuracy is maintained in the subsequent training steps. This is shown in Figure \ref{fig:converge} (b) and (c) where convergence is smooth.

For comparison, we trained three U-NET models that achieved ~ 75\% mIoU, of which two were pruned to 91\% sparsity.  The pruned model in Figure \ref{fig:converge} (b) was trained for sparsity and has double the FLOPs of Figure \ref{fig:converge} (c) that was trained to minimize FLOPs. The FLOPs objective has a significant performance benefit.  

The architecture reduction line plot in Figure \ref{fig:converge} (b) and (c) illustrates our strategy of stepping up $\mu$ and $\lambda$ in the second training step (see Supplementary Material cityscapescrisp.yaml for training parameters). Significantly, this reduction followed by pruning of CNN channels does not affect the cross entropy loss that continues to slowly decrease.  



\begin{figure}[t]
 \centering
    \subfloat[\centering 91.0\% Sparsity]{\scalebox{1}[-1]{\includegraphics[width=3cm]{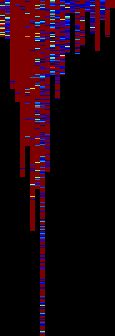} }}%
    \qquad
    \subfloat[\centering 93\% FLOPs reduction]{\scalebox{1}[-1]{\includegraphics[width=3cm]{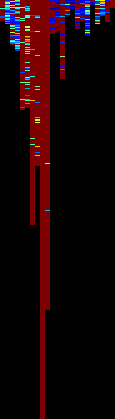} }}%
    \caption{   The relaxation $\sigma(s)$ heatmaps of the U-Net models trained with \textbf{(a)} Sparsity and \textbf{(b)} FLOPs objectives. The columns from left to right are the U-Net subnetworks from model input to output, respectively. From top to bottom are the architecture minimization function output of the subnetwork. The color of each convolutional output is scaled by the relaxation function. Blue indicates a relaxation value of 0 and is to be pruned, while red represents 1 as a channel to be kept. The black background is negative space. See the Supplementary Section for a larger figure.}%
    \label{fig:structure}%
\end{figure}

Figure \ref{fig:structure} illustrates the structure of the pruned U-Net models from Figure \ref{fig:converge} (b) and (c) trained for sparsity (a) and FLOPs (b) after the final step of CRISP. In Figure \ref{fig:converge}, from the left (input) to the right (output) is a column for each of the U-Net subnetworks as defined in Section \ref{Network Augmentation and Pruning}. The original U-Net forms an isosceles triangle. When pruned for both sparsity and FLOPs, the vast majority of the U-Net is carved away.  While the sparsity objective heavily reduced the peak and its adjacent layers of the U-Net in Figure \ref{fig:structure} (a), the FLOPs reduction U-Net is well-pruned in layers toward the edge of the triangle in Figure \ref{fig:structure} (b).  This is intuitive because the U-Net peak has been downsized by 8x. There is a greater reward for the FLOPs objective to remove channels from the outer layers that process and propagate larger resolution data.

Figure \ref{fig:structure} indicates a clear structural different between pruning for Sparsity and FLOPs although they share a common overall pattern.  The pattern of a compressed, rather than symmetric decoder is found in manually-optimized networks similar to the DeeplabV3 architecture \cite{chen2017rethinking}.

\subsection{IoU vs Object Instance Size}\label{IoU & Feature Size}

 \begin{figure*}[ht!]  
  \centering 
\includegraphics[width=\textwidth]{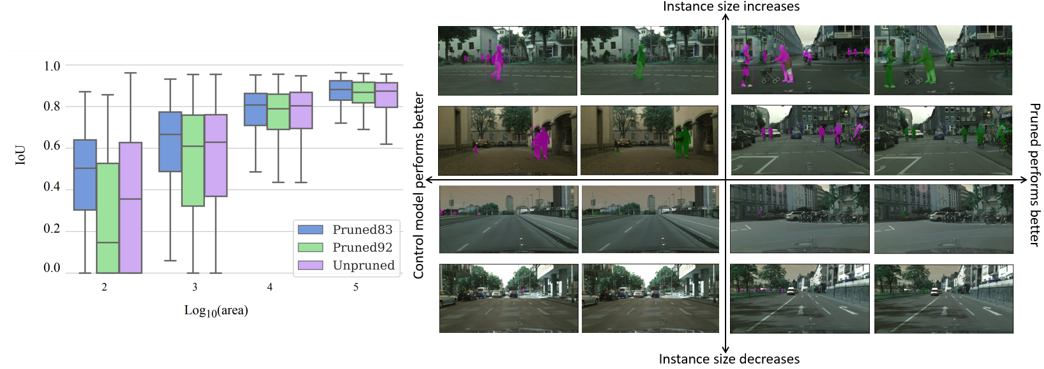}
  \caption{IoU vs Object Instance Size. \textbf{(a)} A comparison of IoU vs sizes of the `human' class instances by pixel. Object sizes are binned by rounding to the closest log base 10 integer. Error bars represent the 1.5*IQR (interquartile range) on either side of the mean. \textbf{(b)} Performance of Pruned and Unpruned (standard) U-Net performance as the instance size changes. We observe little to no difference for the qualitative performance between the models.}
\label{fig:iou_size}
\end{figure*}

 \begin{figure}[ht!]  
  \centering 
\includegraphics[width=.35\textwidth]{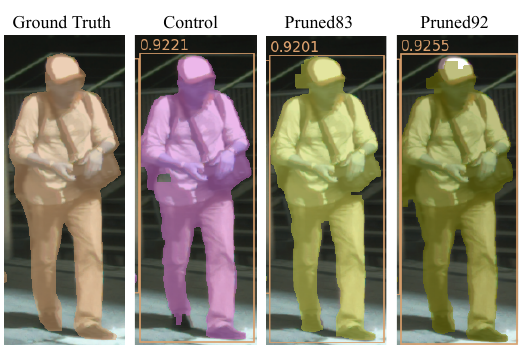}
  \caption{Comparision of ground truth, unpruned network (standard U-Net) and networks pruned to different degrees}
\label{fig:size-changes-with-pruned}
\end{figure}

Finally, we analyze the effect of pruning on the segmentation of objects with varying size. Effectively, we ask, ``does the segmentation of pruned models change when object sizes change?''. We leverage the `human' class of CityScapes with instances that vary greatly in shape and size, and train one standard U-Net and two pruned U-Nets that converge to similar IoUs (~0.74). We present the results of all `human' instances in the validation and test sets. Figure \ref{fig:iou_size} (a) shows the performance of an unpruned U-Net model, and two models that are pruned to 83.16\% (called Pruned83) and 91.92\% (called Pruned92) where these percentages represent the FLOPs removed. We note that all the pruned models maintain performance on all sized objects. A visualization of the effect of pruning on the models' segmentation output are found in \ref{fig:iou_size} (b) and Figure \ref{fig:size-changes-with-pruned}. Figure 
\ref{fig:iou_size} (b) demonstrates variable performances by the Pruned92 and Unpruned models and Figure \ref{fig:size-changes-with-pruned} show the projected segmentation on one zoomed-in example instance. As the pruning progresses, we observe slight changes in the mask output. However, the objects are consistently segmented out. 


\if 
that includes:
\begin{enumerate*}
  \item neural architecture search identifying optimal ML architectures 
  \item model selection
  \item optional pre-train or transfer learning to initialize model
  \item CRISP structural optimization on the target dataset
  \item numeric precision optimization for the target hardware
  \item model computation optimization on the target hardware.
\end{enumerate*}
\fi 

\section{Conclusion}
Whereas large neural networks enable us to uncover increasingly better solutions to complex computer vision problems, a smaller subnetwork within can achieve a similar accuracy. Structured pruning provides a direct increase to computational efficiency within a constrained search space. 


We attained a 96.9\% sparsity with a minimal ($<~1\%$) loss in accuracy with CRISP, on the same ResNet-18 network and CIFAR-10 image classification, an improvement over the current structured state-of-the-art Early-CroP-S algorithm. Sparsity, however, does not represent computational efficiency. Reformulating our objective for FLOPs, we are able to achieve a 93\% FLOPs reduction for CIFAR-10 image classification and CityScapes image segmentation thereby demonstrating CRISPs' extension to more complex computer vision problems. We believe that progressive pruning is beneficial to deploying large models in real-world systems where objects of different sizes show up at test time. We validate performance across objects of different sizes of an important class category and show that the performance does not vary with object size both quantitatively and qualitatively. On ImageNet we estimate 92\% sparsity and 93\% FLOPs reduction with 77\% accuracy, on CityScapes we achieve 75.6\% mIoU, 91.2\% sparsity with 92.9\% FLOPs reduction. 

We hope to shine a light on structured pruning through our efforts. In addition to the direct impact on model efficiency, we laid out a general approach to identify, relax, and prune sub-graphs applicable to a wide variety of neural networks used in computer vision. By combining model accuracy, pruning, and model stiffening into continuous relaxation and optimization regularization we only have one process to perform, which can be integrated into a general model training pipeline. This process has stable convergence to the target maximum sparsity without significant loss of accuracy. 
We believe this is a necessary foundation to provide structured pruning as an automated step of a production model optimization pipeline. Our future work includes the automation of CRISP within a CI/CD pipeline, the inference time speedups of CRISP models in a production setting, the pruning of other commonplace neural network structures, and exploring the relationship between network pruning and dataset covariant shift. 
We believe that this technique can be generally applied to large-scale data center applications and other constrained edge devices as well. Datacenter inference could benefit from reduced energy consumption and server requirements while providing headroom for future capabilities. As is the case with all algorithms, we suggest teams working to incorporate CRISP and other optimizations to be socially considerate and aware of potential target impacts.

\newpage
{\small
\bibliographystyle{ieee_fullname}
\bibliography{main}
}

\newpage
\section*{Supplementary Material} \label{Supplementary}
\begin{figure*}[h!]  
  \centering 
\includegraphics{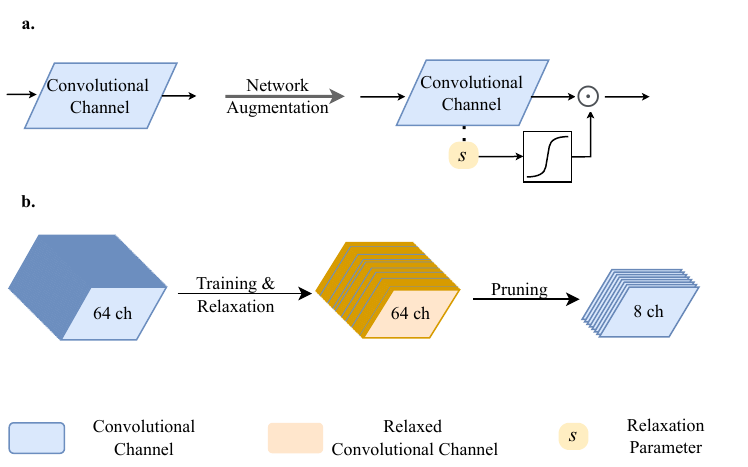}
  \caption{Channel-Level CRISP Diagrams. \textbf{(a)} The CRISP augmentation of an individual convolutional channel. The relaxation parameter is squeezed by a sigmoid function and applied to the convolutional channel's output via element-wise multiplication. \textbf{(b)} A convolutional unit is represented as a stack of 64 convolutional channels. Fifty-six channels become fully relaxed during the training and relaxation step and 8 channels remain after the pruning step.}
\label{fig:ch_diagrams}
\end{figure*}

The supplementary material section contains: 
\begin{enumerate*}
    \item a channel-level visualization of structured pruning (Figure \ref{fig:ch_diagrams}),
    \item the equations describing the parameter and FLOPs computation that is shown in the problem formulation,
    \item a full-page visualization that compare the architectures of unpruned, sparsity objective pruning, and FLOPs objective pruning (Figure \ref{fig:ch_diagrams}).
    \item a table of workflows pushed to Git which can be used to replicate the experiments within the paper 
\end{enumerate*}

Equations 3 and 4 employ Table \ref{tab:objective} for compute model structure $\sigma(s)$ based on the network operator.  These equations were derived from the ptflops and fvcore python libraries.  
\begin{table*}
  \centering
\begin{tabular}{|c|c|c|}
\hline
Operator & Param & FLOPs \\
\hline \hline
ReLu & $p=0$ & $ q = \displaystyle{ \prod T_i }$ where $\displaystyle {T_i[c] = \sum \sigma(s) }$ \\
\hline 
Batch Norm & $p = 2 \cdot \sum \sigma(s)$ & $ q = \displaystyle{ \prod T_i }$ where $\displaystyle {T_i[c] = \sum \sigma(s) }$ \\
\hline 
Sum $\sum$ & $p = 0$ & $ q = \displaystyle{ \prod T_i }$ where $\displaystyle {T_i[c] = \sum \sigma(s) }$\\
\hline 
Product $ v \cdot w $ & $p = 0 $& $ q = \displaystyle{ \prod T_i }$ where $\displaystyle {T_i[c] = \sum \sigma(s) }$\\
\hline 
Concatenation $ \oplus$ & $p = 0 $& $q = 0$\\
\hline 
Convolution $\circledast$ & $p={\displaystyle c_{i} \cdot \sigma_i(s) \cdot c_{o} \cdot \sigma_o(s) \cdot \prod m}$ & $q = c_o \cdot \sigma_o(s) \cdot \prod d_o \cdot ( \prod m \cdot c_i \cdot \sigma_i(s) + 1)$ \\
\hline 
Fully Connected & & \\
\hline
\end{tabular}
  \caption{Selected neural network operators with their trainable parameters, and FLOPs. When no relaxation function is being applied to an operator, $\sum \sigma(s)$ is replaced by the channel length.}
  \label{tab:objective}
\end{table*}

\begin{figure*}
\centering
\begin{subfigure}{.5\textwidth}
  \centering
  \includegraphics[width=6.8cm]{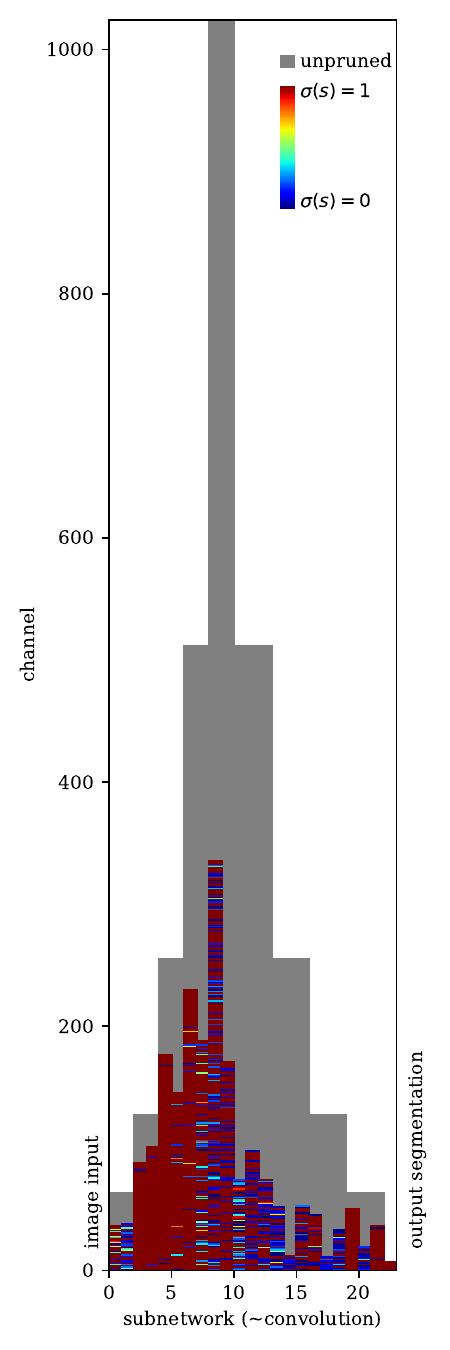}
  \caption{sparsity objective: 75.4\% mIoU, 91.0\% Sparsity, 33.7 GFLOPs}
  \label{fig:sub1}
\end{subfigure}%
\begin{subfigure}{.5\textwidth}
  \centering
  \includegraphics[width=6.8cm]{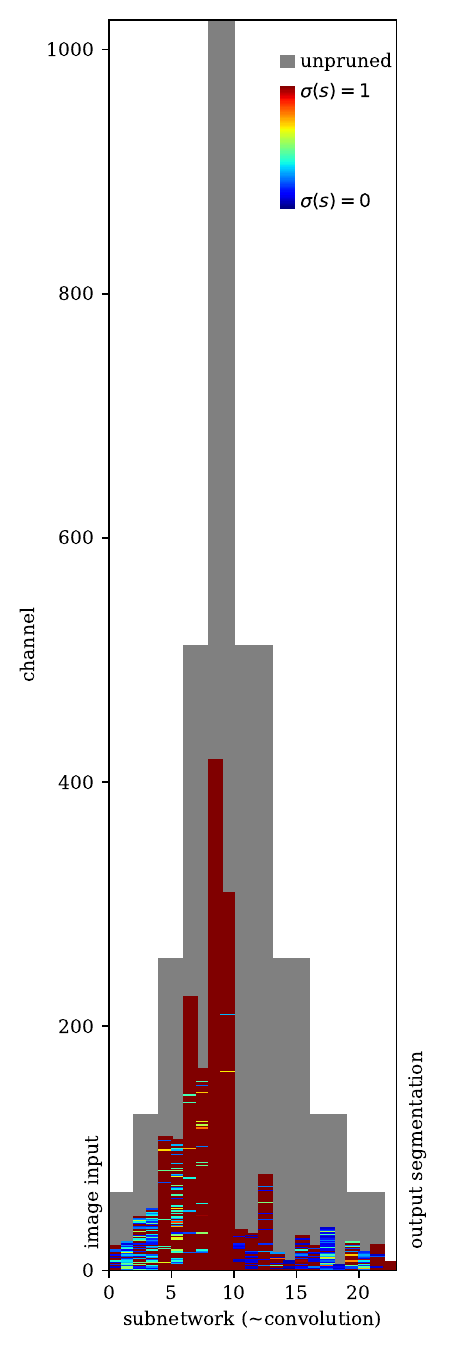}
  \caption{FLOPs objective: 75.6\% mIoU, 91.2\% Sparsity, 15.5 GFLOPs}
  \label{fig:sub2}
\end{subfigure}
\caption{Larger version of Figure 5. The final results of the multi-step pruning of U-Net. The gray space indicates the unpruned U-Net channels. $\sigma(s)$ is the pruning parameter, where channels are pruned when the channel weight is less than the pruning parameter. The pruning parameter lies strictly within (0,1). For multi-step pruning, we generally start with a minimal threshold $(~0.01)$ to avoid pruning channels that could be important later for high accuracy. Blue indicates the lowest values of $\sigma(s)$ and the red indicates the highest value of $\sigma(s)$ when these channels have been pruned. (a) Sparsity objective and (b) FLOPS objective of pruned U-Nets that achieve a similar mIoU. }
\label{fig:test}
\end{figure*}

\begin{table*}
  \centering
\begin{tabular}{|c|c|c|}
\hline
Test & Workflow & Description \\
\hline \hline
Accuracy vs Sparsity & \href{https://github.com/sherlj5352/crisp/blob/main/workflow/resnet18cifar_20221108_142948_ai2.yaml}{resnet18cifar10-dht9m} & Pareto search\\
\hline 
Accuracy vs Sparsity & \href{https://github.com/sherlj5352/crisp/tree/main/workflow/resnet18cifar_20221107_212550_ai2.yaml}{resnet18cifar10-wwjrb} & Pareto search\\
\hline
Accuracy vs Sparsity & \href{https://github.com/sherlj5352/crisp/blob/main/workflow/resnet18cifar.yaml}{resnet18cifar10-8cp4w} & Pareto search\\
\hline
Accuracy vs Sparsity & \href{https://github.com/sherlj5352/crisp/tree/main/workflow/resnet18cifar_20221107_083739_ai2.yaml}{resnet18cifar10-l7892} & Pareto search\\
\hline 
Sparsity vs FLOPs & \href{https://github.com/sherlj5352/crisp/blob/0.1.1/workflow/cityscapes.yaml}{cityscapes} & Unpruned model \\
\hline 
Sparsity vs FLOPs & \href{https://github.com/sherlj5352/crisp/blob/0.1.1/workflow/cityscapescrisp.yaml}{cityscapescrisp} & Sparsity or FLOPs objective models\\
\hline 
 IoU vs Object Instance Size & \href{https://github.com/sherlj5352/crisp/blob/0.1.1/workflow/cityscapes2_control.yaml}{cityscapes2\_control} & Unpruned model \\
\hline 
 IoU vs Object Instance Size & \href{https://github.com/sherlj5352/crisp/blob/0.1.1/workflow/cityscapes2_prune}{cityscapes2\_prune} & Pruned83 and Pruned92 models \\
\hline
\end{tabular}
  \caption{Experiment Argo Workflows}
  \label{tab:objective}
\end{table*}

\end{document}